\documentclass[lettersize,journal]{IEEEtran}

\usepackage{amsmath,amsfonts}
\usepackage{amsthm}
\usepackage{booktabs}
\usepackage{cite}
\usepackage{color}
\usepackage{enumitem}
\usepackage{graphicx}
\usepackage{multirow}
\usepackage{pifont}
\usepackage{tabularx}
\usepackage{subfigure}
\usepackage[table]{xcolor}
\usepackage{balance}
\usepackage[ruled,vlined,linesnumbered]{algorithm2e}
\usepackage{url}
\usepackage{siunitx}

\begin{document}

\title{Stable Multimodal Graph Unlearning via Feature-Dimension Aware Quantile Selection}

\author{Jingjing Zhou, 
Yongshuai Yang, 
Qing Qing, 
Ziqi Xu,~\IEEEmembership{Member,IEEE}
Xikun Zhang, 
Renqiang Luo,~\IEEEmembership{Member,IEEE} 
Ivan Lee,~\IEEEmembership{Senior Member,IEEE}
Feng Xia,~\IEEEmembership{Fellow,IEEE}
\thanks{Jingjing Zhou and Yongshuai Yang are with the School of Information and Electronic Engineering, Zhejiang Gongshang University, Hangzhou, China (e-mail: zhoujingjing@zjgsu.edu.cn, frost.yang@outlook.com).}%
\thanks{Qing Qing and Renqiang Luo are with the College of Computer Science and Technology, Jilin University, Changchun 130012, China (e-mail: qingqing25@mail.jlu.edu.cn, lrenqiang@jlu.edu.cn).}%
\thanks{Ziqi Xu, Xikun Zhang, and Feng Xia are with the School of Computing Technologies, RMIT University, Melbourne, VIC 3000, Australia (e-mail: \{ziqi.xu, xikun.zhang\}@rmit.edu.au, f.xia@ieee.org).}%
\thanks{Ivan Lee is with the School of Computer Science and Information Technology, Adelaide University, Adelaide, SA 5095, Australia (e-mail: ivan.lee@adelaide.edu.au).}%
\thanks{Corresponding author: Xikun Zhang, Renqiang Luo.}
}

\markboth{IEEE Transactions on Multimedia}{Zhou \MakeLowercase{et al.}: Stable Multimodal Graph Unlearning via Feature-Dimension Aware Quantile Selection}

\maketitle

\begin{abstract}
Graph unlearning remains a critical technique for supporting privacy-preserving and sustainable multimodal graph learning. 
However, we observe that existing unlearning strategies tend to apply uniform parameter selection and editing across all graph neural network (GNN) layers, which is especially harmful for multimodal graphs where high-dimensional input projections encode dominant cross-modal knowledge. 
As a result, over-editing these sensitive layers often leads to catastrophic utility degradation after forgetting, undermining both stable learning and effective privacy protection.
To address this gap, we propose FDQ, a Feature-Dimension Aware Quantile framework for multimodal graph unlearning. 
FDQ adaptively identifies high-dimensional input projection layers and applies more conservative, FDQ-guided quantile thresholds when constructing suppression sets, while keeping the underlying importance estimation mechanism unchanged. 
FDQ is seamlessly integrated with diagonal sensitivity-based parameter importance analysis to enable efficient node and edge unlearning under general forget requests.
Through extensive experiments on Ele-Fashion and Goodreads-NC, we demonstrate that FDQ consistently achieves strong utility preservation while maintaining effective forgetting against membership inference attacks. 
Overall, FDQ offers a principled and robust solution for privacy-aware unlearning in high-dimensional multimodal graph systems.
\end{abstract}


\section{Introduction}
\IEEEPARstart{W}{ith} the increasing complexity of network applications such as recommendation systems and social networks, graph-structured data has become a cornerstone for many large-scale online services~\cite{xia2026graph,du2025telling,ren2023graph}. 
The data in these scenarios not only contains structural information but is also increasingly enriched with multimodal content, such as textual descriptions and product images. 
These multimodal graphs provide a more comprehensive characterization of entities and their relationships, offering powerful potential for enhancing user experience and service precision~\cite{du2025Openviewer}. 
In such settings, effective multimodal graph learning often requires going beyond naive feature aggregation by leveraging informative higher-order structural patterns and accounting for heterogeneous modality preferences~\cite{yu2025multi,wang2021dualgnn}.
Concurrently, global data privacy regulations, notably the European Union's General Data Protection Regulation (GDPR), have established a clear ``right to be forgotten,'' mandating that service providers can effectively erase personal data and its influence from their systems upon user request~\cite{regulation2018general,song2026synthetic}.
Therefore, graph unlearning, a technique that enables the efficient removal of specific data points (e.g., nodes or edges) and their influence from a trained graph model, has become crucial for building sustainable, compliant, and privacy-respecting network ecosystems.
\begin{figure}[t]
    \centering
    \includegraphics[width=0.95\linewidth]{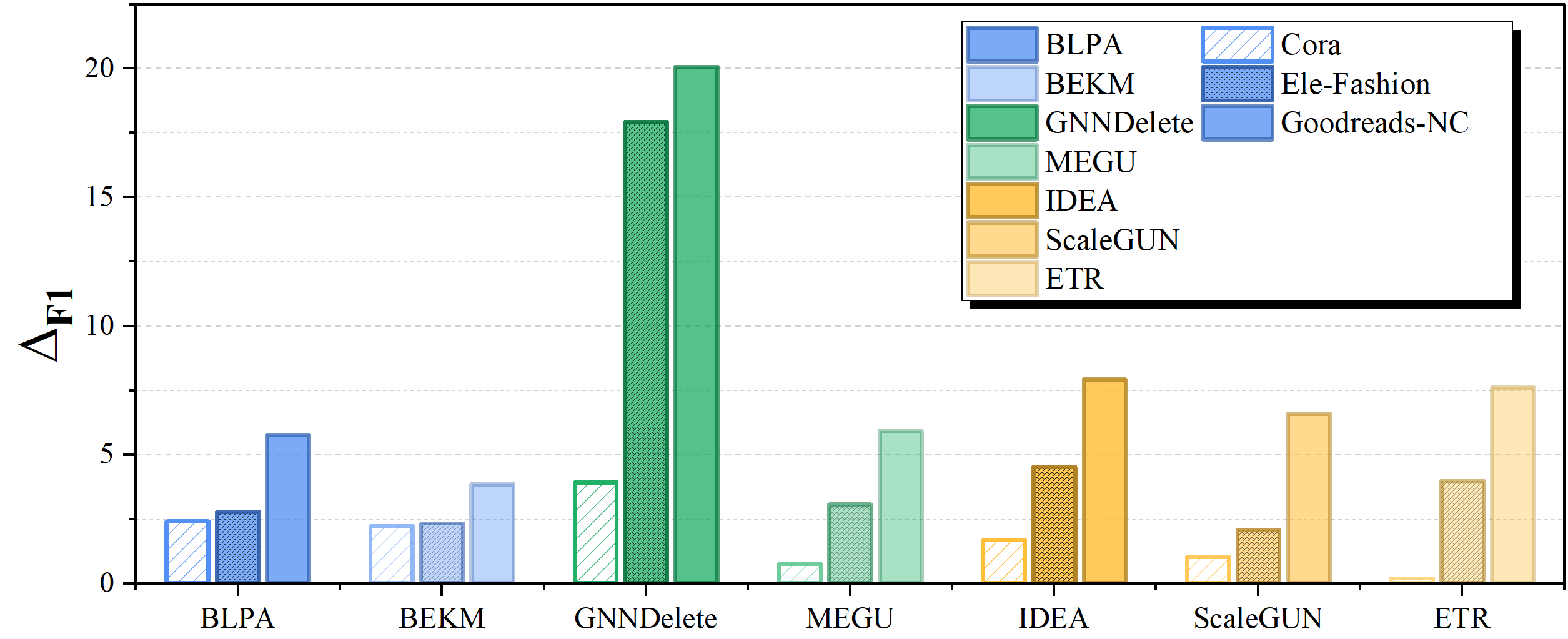}
    \caption{F1 degradation of existing GU methods on conventional and multimodal datasets.}
    \label{fig:mia_result}
\end{figure}

\par However, mainstream graph unlearning methods, such as partition-based GraphEraser~\cite{chen2022graph}, certified unlearning IDEA~\cite{dong2024idea}, mutual-evolution-based MEGU~\cite{li2024towards}, and parameter-editing-based ETR~\cite{yang2025erase}, are primarily designed for graphs with unimodal or low-dimensional node features. 
When applied to graphs with high-dimensional multimodal features, these methods often lead to severe utility degradation on the retained data after unlearning. 

\par This issue stems from the excessive modification of first-layer projection matrices in graph neural networks (GNNs). Multimodal GNNs are structurally imbalanced: input projection layers that process high-dimensional features (e.g., CLIP~\cite{radford2021learning} or ImageBind~\cite{girdhar2023imagebind}) dominate the parameter space and encode substantial cross-modal knowledge. 
Existing methods adopt uniform parameter selection and updating strategies, resulting in indiscriminate edits to these sensitive layers and damaging pre-trained representations and generalization. 
Consequently, they either incur utility collapse or weaken forgetting to preserve performance. 
This highlights a key gap: current methods lack a unified framework for stable and effective unlearning in high-dimensional multimodal graphs.

\par To address this, we propose \textbf{FDQ}, a \textbf{F}eature \textbf{D}imension-aware \textbf{Q}uantile framework for multimodal graph unlearning. 
Rather than treating all layers uniformly, FDQ adapts update strength based on feature dimension and layer role, protecting wide input projections while allowing deeper layers to remain editable.

\par Concretely, FDQ identifies high-dimensional input layers and applies conservative layer-wise quantile thresholds when constructing suppression sets, without altering the underlying importance estimation. 
This design integrates seamlessly with efficient diagonal importance estimation, supporting both node and edge unlearning under general forget requests. 
Overall, FDQ mitigates over-editing in input layers while enabling effective forgetting through selective updates in deeper layers.

The main contributions of this work are summarized as follows:
\begin{itemize}[leftmargin=0.5cm]
    \item We identify a key limitation of existing graph unlearning methods: when applied to multimodal graphs, they suffer from severe utility degradation due to parameter imbalance and over-editing of input projection layers.
    \item We propose FDQ, a feature-dimension-aware quantile framework that adapts parameter selection thresholds, enabling stable unlearning while preserving essential multimodal knowledge. It supports both node and edge unlearning without reconstructing raw data.
    \item Extensive experiments on real-world datasets show that FDQ consistently outperforms state-of-the-art methods in both utility preservation and privacy protection.
\end{itemize}

\section{Related Work}

\subsection{Graph Unlearning}
\par Driven by privacy regulations such as the GDPR's ``right to be forgotten'', graph unlearning has emerged as a key technique to remove the influence of specific data points (nodes, edges, or features) from a trained GNNs without costly retraining~\cite{regulation2018general}. 
Existing methods can be broadly categorized into three paradigms.

\par Partition-based methods split the graph into approximately independent shards, train a sub-model on each, and retrain only affected shards upon unlearning requests. 
GraphEraser~\cite{chen2022graph} adapts the Sharded, Isolated, Sliced, and Aggregated (SISA) framework~\cite{bourtoule2021machine} to graphs using strategies such as balanced label propagation. While efficient, it is designed for static, transductive settings. Subsequent work like GUIDE~\cite{wang2023inductive} extends this approach to dynamic, inductive scenarios. 
However, the performance remains constrained by partition quality, and the ``divide-and-conquer'' paradigm struggles to preserve global, high-dimensional representations in multimodal graphs.

\par Learning-based methods design specialized objectives to endow models with inherent unlearning capabilities. 
GNNDelete~\cite{cheng2023gnndelete} applies gradient-ascent on the forget set with knowledge distillation on the retain set, while 
MEGU~\cite{li2024towards} proposes a mutual evolution paradigm that unifies training and unlearning via a bi-level optimization. 
Although achieving a good utility-forgetting trade-off, these general frameworks typically apply uniform update rules across all layers, which can harm high-dimensional input layers encoding pre-trained multimodal knowledge.

\par Influence function-based methods estimate and negate the effect of training data using influence functions or related techniques. 
IDEA~\cite{dong2024idea} provides a flexible framework for certified graph unlearning. 
ScaleGUN~\cite{yi2025scalable} improves scalability via randomized blockwise influence estimation. 
ETR~\cite{yang2025erase} proposes a training-free two-stage parameter editing approach. Despite strong theoretical grounding, these methods often use a global, uniform thresholds for parameter selection or editing, ignoring the structural parameter imbalance in multimodal GNNs, where the input projection layers contain a dominant fraction of parameters. 
This oversight makes them prone to catastrophically degrading model utility when applied to high-dimensional multimodal graphs.

\par In summary, despite progress in efficiency, certification, and generality, existing graph unlearning methods commonly rely on uniform rules for partitioning, updating, or parameter editing. 
Such designs are brittle for multimodal graphs, where high-dimensional features make the input projection layers parameter-dominant and prone to over-editing. 
FDQ addresses this limitation via a feature-dimension-aware quantile mechanism for stable unlearning on multimodal graphs.

\subsection{Multimodal Graph Learning}
\par Multimodal Graph Learning (MMGL) integrates information from multiple modalities (e.g., text, images) associated with graph entities to learn richer representations~\cite{peng2024learning}. 
In applications like social networks, e-commerce, and healthcare, nodes naturally carry complementary textual and visual content~\cite{chen2021heterogeneous,dong2025multimodal}. 
Effectively harnessing these modalities is crucial for improving performance on downstream tasks such as node classification and link prediction~\cite{ektefaie2023multimodal,he2025unigraph2,fang2025graphgpt,jian2026hierarchy}. 
Early methods focused on graph topology and simple features. The rise of large pre-trained language models (LMs) shifted significant attention to text-attributed graphs and GNN-LM architectures~\cite{jin2024large,yan2023comprehensive,shehzad2026graph}.
However, visual information provides unique semantic cues that text and structure alone cannot capture, such as stylistic similarities in products or artistic works~\cite{wei2024gita}.
This underscores the need for models that jointly exploit textual, visual, and structural information~\cite{zhang2025vidr}.

\par To standardize evaluation in this emerging field, Zhu et al. introduced MM-GRAPH, the first comprehensive benchmark incorporating both high-resolution visual and rich textual information for graph learning~\cite{zhu2025mosaic}. 
MM-GRAPH comprises seven diverse datasets spanning tasks like node classification, link prediction, and knowledge graph completion, and systematically studies multimodal feature encoding strategies.
It compares aligned encoders (e.g. CLIP~\cite{radford2021learning}, ImageBind~\cite{girdhar2023imagebind}), which project different modalities into a unified aligned embedding space, with non-aligned encoder combinations (e.g., T5~\cite{raffel2020exploring}+ViT~\cite{dosovitskiy2020image}).
Results show that cross-modal alignment consistently improves performance, and that multimodal models outperform single-modality ones, highlighting the value of visual information.

\par While MMGL has advanced in multimodal feature encoding and fusion~\cite{tang2026rmtrans}, most studies overlook lifecycle requirements such as selective data removal~\cite{zhang2022cglb}. 
Recent work on multimodal unlearning~\cite{cheng2024multidelete,sinha2025multi} focus on settings distinct from multimodal graph unlearning. 
In multimodal graphs, high-dimensional features make input projection layers parameter-dominant, so naive unlearning can severely disrupt pre-trained cross-modal representations. 
This gap motivates feature-dimension-aware unlearning mechanisms, as exemplified by our FDQ framework.

\section{Preliminaries}
\subsection{Notations}
\par Unless stated otherwise, the following conventions are used for mathematical notions: sans-serif uppercase letters (e.g., $\mathcal{V}$, $\mathcal{E}$) denote sets, bold uppercase letters (e.g., $\mathbf{A}$, $\mathbf{X}$) denote matrices, and bold lowercase letters (e.g., $\mathbf{x}_v$) denote vectors. 
A multimodal graph is defined as $\mathcal{G} = (\mathcal{V}, \mathcal{E}, \mathbf{X})$, where $\mathcal{V} = \{v_1, v_2, ..., v_n\}$ is the set of $n$ nodes, and $\mathcal{E} \subseteq \mathcal{V} \times \mathcal{V}$ is the set of edges. 
The corresponding adjacency matrix is $\mathbf{A} \in \{0, 1\}^{n \times n}$, where $\mathbf{A}_{ij} = 1$ if $(v_i, v_j) \in \mathcal{E}$ and $0$ otherwise. 
$\mathbf{X} \in \mathbb{R}^{n \times d}$ is the multimodal feature matrix of nodes, where the $v$-th row $\mathbf{x}_v \in \mathbb{R}^{d}$ is the high-dimensional feature vector for node $v$. 
The feature dimension $d$ is typically large (e.g., 1024, 1536, 2048) and is produced by pre-trained multimodal encoders (e.g., CLIP, ImageBind). 
The class label for a node is denoted as $y_v \in \mathcal{Y}$.

\subsection{Graph Unlearning}
\par Graph unlearning extends machine unlearning to graph-structured data, aiming to remove the influence of specified training samples (e.g., nodes or edges) from a trained GNN without costly retraining. 
It operates at the model level: after an unlearning request, the updated parameters should behave as if the forget set had never been used in training. 
This capability is important in real deployments where deletion requests may arise during a model's lifecycle.

\par Formally, let $f_{\theta}$ be a GNN trained on the full training set $\mathcal{D}_{train}\subseteq \mathcal{V}$. 
An unlearning request specifies a forget set $\mathcal{D}_f\subseteq \mathcal{D}_{train}$, and the retain set is defined as $\mathcal{D}_r=\mathcal{D}_{train}\setminus \mathcal{D}_f$. 
We evaluate the model on a separate test set $\mathcal{D}_{test}$, and denote by $f_{\theta^*}$ the ideal model retrained from scratch solely on $\mathcal{D}_r$. 
Given $(f_{\theta},\mathcal{D}_f)$, an unlearning algorithm outputs an updated model $f_{\theta'}$ that satisfies
\begin{equation}
f_{\theta} \xrightarrow[]{\mathcal{D}_f} f_{\theta'} \approx f_{\theta^*}.
\label{eq:unlearning_def}
\end{equation}
\par This approximation should preserve performance on retained data while removing the influence of $\mathcal{D}_f$. 
In practice, it is evaluated via downstream performance and privacy leakage on the forget set. 
An effective method should also be significantly faster than retraining.  
In this paper, we consider two request types: node unlearning, which removes target nodes and their incident edges, and edge unlearning, which only removes specific edges. 
We evaluate forgetting using membership inference attacks (MIA) under the MIA-Graph threat model~\cite{olatunji2021membership}, which assess residual membership signals of the forget set, and test robustness with poisoning attacks (PA) to verify the removal of poisoned edge influence.

\subsection{Multimodal Graph Neural Network}
\par Multimodal GNNs operate on graphs where each node carries multimodal content (e.g., text and images). 
Pre-trained encoders map raw modalities into high-dimensional features $\mathbf{x}_v\in\mathbb{R}^{d}$, forming a feature matrix $\mathbf{X}\in\mathbb{R}^{n\times d}$. 
A common design is to first apply an input projection that maps $\mathbf{x}_v$ into a hidden space of dimension $h$:
\begin{equation}
\mathbf{h}_v^{(0)} = \sigma\!\left(W_{\text{in}}\mathbf{x}_v + \mathbf{b}_{\text{in}}\right), \qquad W_{\text{in}}\in\mathbb{R}^{h\times d},
\label{eq:input_projection}
\end{equation}
where $\sigma(\cdot)$ is a non-linear activation. 
After the input projection, the GNN applies $K$ message-passing layers. For $k=1,\dots,K$:
\begin{equation}
\mathbf{h}_v^{(k)} = \phi^{(k)}\left( \mathbf{h}_v^{(k-1)},\; \psi^{(k)}\left( \{ \mathbf{h}_u^{(k-1)} : u\in \mathcal{N}(v) \} \right) \right),
\end{equation}
where $\mathcal{N}(v)$ denotes the neighborhood of node $v$,
and $\phi^{(k)}$ and $\psi^{(k)}$ instantiate the update and aggregation operations of a specific GNN architecture (e.g., convolutional or attention-based variants).
\par The final embedding $\mathbf{h}_v^{(K)}$ is fed into a task head (e.g., a linear classifier) to obtain $\hat{y}_v$, and we denote the node-wise output as $f_{\theta}(\mathbf{X},\mathbf{A})_v$. 
The above formulation is architecture-agnostic and covers common GNN variants.
\begin{figure*}[t]
    \centering
    \includegraphics[width=0.95\textwidth]{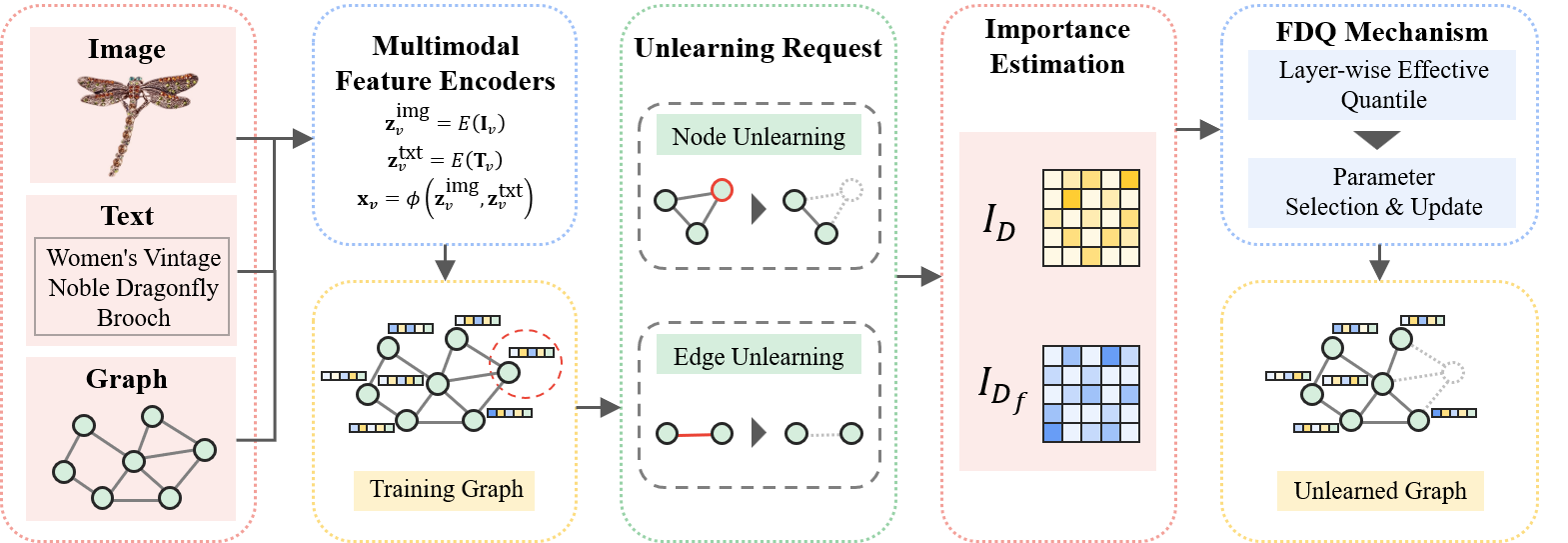}
    \caption{Overall pipeline of FDQ for multimodal graph unlearning.}
    \label{fig:framework}
\end{figure*}
\section{FDQ for Multimodal Graph Unlearning}

\begin{algorithm}[t]
    \caption{FDQ: Multimodal Node Unlearning}
    \label{alg:fdq_node}
    \SetAlgoLined
    \KwIn{multimodal graph $\mathcal{G}=(\mathcal{V},\mathbf{A},\mathbf{X})$, trained multimodal GNN $f_{\theta}$, training set $\mathcal{D}$, forget node set $F$, hyperparameters $h,k,\rho,k_{\min},\tau,\gamma$}
    \KwOut{Unlearned parameters $\theta'$}
    
    Construct neighbor set $\mathcal{D}_{\text{nbr}}=\mathcal{N}_h(F)\setminus F$\;
    Compute diagonal FIMs $\mathbf{F}_{\mathcal{D}}$, $\mathbf{F}_{\mathcal{D}_f}$, $\mathbf{F}_{\mathcal{D}_{\text{nbr}}}$ using Eq.~\eqref{eq:fim_diag_approx}\;
    
    \For{each parameter tensor $W^{(j)}$ in $\theta$}{
        Compute suppression scores $b_1^{(j)}, b_2^{(j)}$ by Eq.~\eqref{eq:scores_node}\;
        Compute $\alpha^{(j)}$ by Eq.~\eqref{eq:alpha_def} and $k_{\text{eff}}^{(j)}=\alpha^{(j)}\cdot k$ by Eq.~\eqref{eq:k_eff_def}\;
        Set thresholds $t_1^{(j)}=Q_{k_{\text{eff}}^{(j)}}(b_1^{(j)})$ and $t_2^{(j)}=Q_{k_{\text{eff}}^{(j)}}(b_2^{(j)})$\;
        Select $\Omega^{(j)}=\{p \mid b_1^{(j)}[p]\le t_1^{(j)} \ \lor\ b_2^{(j)}[p]\le t_2^{(j)}\}$ by Eq.~\eqref{eq:omega_node}\;
        
        \For{each $p \in \Omega^{(j)}$}{
            Determine $(b,t)\leftarrow(b_1^{(j)}[p],t_1^{(j)})$ if $b_1^{(j)}[p]\le t_1^{(j)}$, else $(b,t)\leftarrow(b_2^{(j)}[p],t_2^{(j)})$\;
            Update $\theta'_j[p]\leftarrow \theta_j[p]\cdot \min\!\left(\frac{b}{t},\gamma\right)$ by Eq.~\eqref{eq:update_node}\;
        }
    }
    \Return $\theta'$\;
\end{algorithm}

\par In this section, we present Feature-Dimension Aware Quantile (FDQ) for multimodal graph unlearning. 
Our main technical contribution, the FDQ mechanism, addresses a key limitation of parameter-editing unlearning on multimodal graphs: uniform quantile thresholds across layers tend to over-select parameters in wide input projection matrices, degrading pre-trained cross-modal representations. 
FDQ mitigates this by adaptively tightening effective quantiles for high-dimensional input layers while while allowing more selective updates in deeper layers under the same importance signals. 
To define forgetting targets, we adopt diagonal FIM approximation to score parameter sensitivity to retained data versus data affected by the forget request (including neighbor contexts in node unlearning), and then apply FDQ when converting those scores into suppression sets. 
The resulting pipeline supports both node and edge unlearning by pairing diagonal FIM scoring with FDQ-guided, layer-wise thresholding tailored to high-dimensional multimodal inputs.

\subsection{The FDQ Mechanism: Feature-Dimension Aware Quantile}

\label{subsec:fdq}

\par Multimodal graph datasets often have extremely high-dimensional input features. 
The first-layer projection matrices, which map these features into the latent space, contain a large share of model parameters and are crucial for preserving pre-trained multimodal knowledge. 
Uniform parameter selection during unlearning disproportionately alters these layers, causing severe utility degradation. 
FDQ mitigates this by applying more conservative updates to these sensitive parameters.

\par We begin by partitioning the model parameters. Let $\theta = \{W^{(1)}, W^{(2)}, \dots, W^{(L)}\}$ denote all trainable weight matrices in the GNN. 
We define the set of input-projection layers as:
\begin{equation}
\theta_{\text{in}} = \{ W^{(\ell)} \in \mathbb{R}^{m_\ell \times d} \},
\end{equation}
where $d$ is the input feature dimension. 
The complementary set $\theta_{\text{deep}} = \theta \setminus \theta_{\text{in}}$ constitutes the deeper layers. 
Intuitively, $\theta_{\text{in}}$ contains parameters that directly process the raw, high-dimensional multimodal features.

\par Instead of applying a global quantile $k$ to all parameters, FDQ assigns a layer-wise effective quantile $k_{\text{eff}}^{(\ell)}$. 
This is achieved by first defining a layer-specific scaling factor $\alpha^{(\ell)}$:
\begin{equation}
\label{eq:alpha_def}
\alpha^{(\ell)} =
\begin{cases}
\max(\rho, \; k_{\min}/k), & \text{if } d \ge \tau \ \land\ W^{(\ell)} \in \theta_{\text{in}}, \\[4pt]
1, & \text{otherwise},
\end{cases}
\end{equation}
where $\rho \in (0, 1)$ is the tightening ratio, $k_{\min}$ is a lower bound, and $\tau$ is a feature dimension threshold. 
The effective quantile is then obtained by scaling the base quantile $k$:
\begin{equation}
\label{eq:k_eff_def}
k_{\text{eff}}^{(\ell)} = \alpha^{(\ell)} \cdot k.
\end{equation}
\par In ths way, for high-dimensional inputs ($d \ge \tau$), the selection threshold for parameters in $\theta_{\text{in}}$ is tightened (at least by a factor of $\rho$), promoting sparser updates, while deeper layers remain unaffected.

\par The core effect of FDQ is to reduce the number of parameters selected for modification in $\theta_{\text{in}}$. 
Consider a per-parameter suppression score $b^{(j)}$ (see Sec.~\ref{subsec:integrated_fdq}), which quantifies how specialized a parameter is to the data intended for forgetting.
The selected parameter set under a standard quantile rule is:
\begin{equation}
\label{eq:omega_standard}
\Omega^{(j)}(k) = \{ p \mid b^{(j)}[p] \le Q_k(b^{(j)}) \},
\end{equation}
where $Q_k(\cdot)$ computes the $k$-th quantile. With FDQ, the selection set becomes:
\begin{equation}
\label{eq:omega_fdq}
\Omega^{(j)}_{\text{FDQ}} = \{ p \mid b^{(j)}[p] \le Q_{k_{\text{eff}}^{(j)}}(b^{(j)}) \}.
\end{equation}
\par Since $k_{\text{eff}}^{(j)} \le k$ for $W^{(j)} \in \theta_{\text{in}}$, it follows that $|\Omega^{(j)}_{\text{FDQ}}| \le |\Omega^{(j)}(k)|$, enforcing sparser modifications on the critical input-projection layers.

\par The necessity of the FDQ mechanism is rooted in the parameter distribution of GNNs processing high-dimensional features. 

The parameters in $\theta_{\text{in}}$ often constitute a dominant fraction of the model's total capacity. 
This can be quantified as:
\begin{equation}
\label{eq:param_ratio}
R_{\text{in}} = \frac{ |\theta_{\text{in}}| }{ |\theta| } = \frac{d \cdot \sum_{\ell \in \mathcal{L}_{\text{in}}} m_\ell}{d \cdot \sum_{\ell \in \mathcal{L}_{\text{in}}} m_\ell + C } \propto \frac{d}{d + h_{\text{hid}}},
\end{equation}
where $h_{\text{hid}}$ is the hidden dimension, $\mathcal{L}_{\text{in}}$ is the index set of input layers, and $C$ represents the total number of parameters in deeper layers (roughly on the order of $h_{\text{hid}}^2$).
When the input feature dimension is significantly larger than the hidden dimension ($d \gg h_{\text{hid}}$), the ratio $R_{\text{in}}$ becomes substantial. 
Consequently, applying a uniform quantile $k$ for parameter selection will result in a disproportionally large number of parameters being modified in $\theta_{\text{in}}$ compared to $\theta_{\text{deep}}$. 
Given that $\theta_{\text{in}}$ encodes crucial, pre-trained multimodal representations, such dense and aggressive modifications easily destabilize the model and lead to catastrophic utility degradation. 
FDQ explicitly compensates for this structural imbalance by applying a more conservative (tighter) update policy specifically to $\theta_{\text{in}}$, as defined in Eq.~(\ref{eq:alpha_def}).

\subsection{FIM-based Parameter Importance Analysis}

\par The FIM serves as a fundamental tool for quantifying the sensitivity of model parameters with respect to the data distribution~\cite{kirkpatrick2017overcoming}. 
Formally, for a probabilistic model $p(y|x,\theta)$, the FIM is defined as the expectation of the outer product of the score function:
\begin{equation}
    \label{eq:fim_definition}
    \mathbf{F}(\theta) = \mathbb{E}_{x, y} \left[ \nabla_{\theta} \log p(y|x,\theta) \; \nabla_{\theta} \log p(y|x,\theta)^{\top} \right],
\end{equation}
which is also equivalent to the negative expected Hessian of the log-likelihood. 
It captures the amount of information that an observable random variable carries about the parameters, providing a principled measure of parameter importance.

\par However, computing the full FIM is computationally prohibitive for large-scale GNNs due to its quadratic size in the number of parameters. 
Following established practice~\cite{kirkpatrick2017overcoming}, we adopt an efficient FIM-based importance approximation.
For a given node set $\mathcal{D}$ (e.g., the training set $\mathcal{D}_{train}$, the forget set $\mathcal{D}_f$), the importance for each parameter is estimated by the average squared gradient:
\begin{equation}
    I_{\mathcal{D}}(\theta) \approx \frac{1}{|\mathcal{D}|} 
    \sum_{v \in \mathcal{D}} 
    \left( \nabla_{\theta} \, \mathcal{L} \big( f_{\theta}(\mathbf{X}, \mathbf{A})_v, y_v \big) \right)^2,
    \label{eq:fim_diag_approx}
\end{equation}
where $\mathcal{L}$ is the task-specific loss function (e.g., cross-entropy for node classification). This approximation yields a per-parameter importance score, where a larger value indicates greater sensitivity of that parameter to the data in $\mathcal{D}$.

\par In our framework, we compute $I_{\mathcal{D}}(\theta)$ on the retained training set to estimate knowledge that should be preserved, and $I_{\mathcal{D}_f}(\theta)$ (or $I_{\mathcal{D}_{\text{infl}}}(\theta)$ for edge unlearning) on the data to be forgotten, in order to identify and suppress the corresponding specialized knowledge.

\subsection{Integrated Unlearning Framework with FDQ}
\label{subsec:integrated_fdq}

\par The FDQ unlearning framework integrates the FDQ mechanism with FIM-based parameter importance. 
Given a forget request, it executes a unified three-step pipeline: (1) compute relevant Fisher matrices, (2) construct per-parameter suppression scores, and (3) select and scale parameters using the FDQ-guided quantile.

\par Node unlearning with neighbor influence: for a forget node set $V_f \subseteq \mathcal{V}$, we define $\mathcal{D}_f = V_f$ and its neighbor set $\mathcal{D}_{\text{nbr}} = \mathcal{N}_h(V_f) \setminus V_f$ to capture message-passing effects.
We compute three FIM-based importance metrics: $I_{\mathcal{D}}(\theta)$ (training set), $I_{\mathcal{D}_f}(\theta)$ (forget set), and $I_{\mathcal{D}_{\text{nbr}}}(\theta)$ (neighbor set).

\par Two element-wise suppression scores are constructed for each parameter tensor $W^{(j)}$:
\begin{equation}
\label{eq:scores_node}
b_1^{(j)} = \frac{I_{\mathcal{D}}^{(j)}(\theta)}{I_{\mathcal{D}_f}^{(j)}(\theta)}, \qquad
b_2^{(j)} = \frac{\big(I_{\mathcal{D}}^{(j)}(\theta)\big)^2}{I_{\mathcal{D}_f}^{(j)}(\theta) \cdot I_{\mathcal{D}_{\text{nbr}}}^{(j)}(\theta)}.
\end{equation}
\par A smaller $b_1^{(j)}$ indicates stronger specialization to the forget set, while $b_2^{(j)}$ further penalizes parameters important to both $V_f$ and its neighbors.

\par For each parameter $W^{(j)}$, we apply the FDQ rule (Eq.~\ref{eq:k_eff_def}) to obtain its effective quantile $k_{\text{eff}}^{(j)}$. 
The parameters selected for suppression are:
\begin{equation}
\label{eq:omega_node}
\Omega^{(j)} = \left\{ p \mid b_1^{(j)}[p] \le t_1^{(j)} \ \lor\ b_2^{(j)}[p] \le t_2^{(j)} \right\},
\end{equation}
where $t_1^{(j)}=Q_{k_{\text{eff}}^{(j)}}(b_1^{(j)})$ and $t_2^{(j)}=Q_{k_{\text{eff}}^{(j)}}(b_2^{(j)})$ are the FDQ-adjusted thresholds. 
Finally, each selected parameter is dampened by its score-to-threshold ratio:
\begin{equation}
\label{eq:update_node}
\theta_j[p] \leftarrow \theta_j[p] \cdot \min\left( \frac{b^{(j)}[p]}{t^{(j)}[p]},\; \gamma \right), \quad p \in \Omega^{(j)},
\end{equation}
where $b^{(j)}[p]$ and $t^{(j)}[p]$ denote the score and threshold associated with the criterion by which $p$ is selected (i.e., via $b_1$ or $b_2$), and $\gamma$ is the maximum scaling factor used to prevent excessive suppression.

\par Edge unlearning via influenced-set approximation: for a forget edge set $E_f \subseteq \mathcal{E}$, we approximate its effect via the influenced node set $\mathcal{D}_{\text{infl}} = \mathcal{V}_f \cup \mathcal{N}_h(\mathcal{V}_f)$, where $\mathcal{V}_f$ contains all endpoints of edges in $E_f$. 
We compute $I_{\mathcal{D}}(\theta)$ and $I_{\mathcal{D}_{\text{infl}}}(\theta)$, and construct the suppression score:
\begin{equation}
b^{(j)} = \frac{I_{\mathcal{D}}^{(j)}(\theta)}{I_{\mathcal{D}_{\text{infl}}}^{(j)}(\theta)}.
\end{equation}
\par The same selection-and-scaling procedure applies, except that only a single score $b^{(j)}$ is used. 
The FDQ-adjusted threshold is $t^{(j)} = Q_{k_{\text{eff}}^{(j)}}(b^{(j)})$, defining the selected set $\Omega^{(j)} = \{p \mid b^{(j)}[p] \le t^{(j)}\}$. 
Parameters are then updated via Eq.~(\ref{eq:update_node}) (with $b^{(j)}$ and $t^{(j)}$). 
This suppresses parameters most sensitive to the subgraph affected by the edge removal, while FDQ protects the input-projection layers.

\par In summary, the FDQ framework provides a principled approach for graph unlearning. 
Its core innovation, the FDQ mechanism (Sec.~\ref{subsec:fdq}), adaptively tightens parameter selection for high-dimensional input layers, preventing catastrophic utility drop. 
This mechanism is seamlessly integrated into FIM-based node and edge unlearning pipelines, which identify forget-specific parameters via Fisher ratios and suppress them via FDQ-guided scaling. 
The result is a targeted, efficient, and stable unlearning method suitable for multimodal graphs.

\section{Experiments}
\subsection{Datasets}
\par We evaluate our framework on two multimodal node classification datasets from the MM-GRAPH benchmark~\cite{zhu2025mosaic}. 
Both datasets contain rich textual and visual features associated with nodes, enabling a comprehensive evaluation of multimodal graph unlearning methods. 
Table~\ref{tab:datasets} summarizes their statistics.

\begin{table}[t]
    \centering
    \caption{Statistics of multimodal node classification datasets.}
    \label{tab:datasets}
    \renewcommand{\arraystretch}{1.2}
    \setlength{\tabcolsep}{8pt}
    \begin{tabular}{lll}
        \toprule
        \textbf{Feature} & \textbf{Ele-Fashion} & \textbf{Goodreads-NC} \\ \midrule
        \# Nodes & $97,766$ & $685,294$ \\
        \# Edges & $199,602$ & $7,235,084$ \\
        \# Labels & $11$ & $11$ \\
        Task & Node Classification & Node Classification \\
        Text Features & Fashion titles & Book descriptions \\
        Visual Features & Fashion images & Book covers \\ 
        \bottomrule
    \end{tabular}
\end{table}

\par \textbf{Ele-Fashion} is derived from the Amazon-Fashion corpus~\cite{hou2024bridging,ni2019justifying}, where nodes represent products and edges denote co-purchasing relations. 
Textual features are product titles, and visual features are high-resolution images. 
After preprocessing, it contains 11 valid categories (e.g., shoes, jewelry, dresses), as one original label ID has no samples, resulting in a medium-scale dataset with strong homophily.

\par \textbf{Goodreads-NC} is built from the Goodreads Book Graph~\cite{wan2018item,wan2019fine}, where nodes correspond to books and edges capture user co-preference. 
Textual features are book descriptions, and visual features are cover images, with nodes lacking images removed. 
It contains 11 categories (e.g., History, Children, Comics), forming a large-scale dataset with moderate homophily.

\subsection{Feature Encoders}
\par Following the MM-GRAPH benchmark proposed by Zhu et al.~\cite{zhu2025mosaic}, we adopt a set of state-of-the-art (SOTA) text and visual encoders to extract multimodal node features. 
This ensures consistency with prior work and allows us to fairly evaluate the role of modality alignment and feature representation in graph unlearning.

\par \textbf{Text encoders.} We employ CLIP~\cite{radford2021learning}, T5~\cite{raffel2020exploring}, and ImageBind~\cite{girdhar2023imagebind} as text encoders. 
T5 provides a strong baseline for text representation, while CLIP enables semantic alignment between text and images. ImageBind further extends this capability by embedding multiple modalities into a unified space, benefiting multimodal tasks.

\par \textbf{Visual encoders.} For visual features, we consider CLIP~\cite{radford2021learning}, ViT~\cite{dosovitskiy2020image}, ImageBind~\cite{girdhar2023imagebind}, and DINOv2~\cite{oquab2023dinov2}. 
ViT represents a supervised transformer-based vision model, while DINOv2 provides self-supervised visual embeddings that are robust across domains. 
CLIP and ImageBind are again chosen for their ability to align visual features with text, facilitating cross-modal representation learning.

\subsection{Baselines}
\par To evaluate the performance of FIM-FDQ, we compare it against three major categories of SOTA graph unlearning methods, encompassing seven representative algorithms. 

\par Partition-based Methods decompose the original graph into multiple, approximately independent shards, training a sub-model on each. 
Unlearning is achieved by retraining only the affected shards, thereby avoiding full-model retraining.

\begin{itemize} [leftmargin=0.5cm]
        \item \textbf{GraphEraser-BLPA}~\cite{chen2022graph} uses Balanced Label Propagation to partition the graph into balanced, highly connected shards that promote intra-shard homophily.
        \item \textbf{GraphEraser-BEKM}~\cite{chen2022graph} adopts Balanced $k$-Means to cluster nodes by feature similarity, forming partitions with homogeneous node attributes.
\end{itemize}

\par Learning-based Methods design specialized training objectives or frameworks to endow the model with the ability to forget, often by directly incorporating unlearning signals into the optimization.

\begin{itemize} [leftmargin=0.5cm]
    \item \textbf{GNNDelete}~\cite{cheng2023gnndelete} proposes a gradient-ascent-based strategy, degrading performance on the forget set via a negative loss while preserving knowledge on the retain set through distillation.
    \item \textbf{MEGU}~\cite{li2024towards} introduces a mutual evolution paradigm that unifies training and unlearning via bi-level optimization. 
    It co-trains a predictor and an unlearner, jointly optimizing accuracy on retained data and forgetting quality across node, edge, and feature settings.

\end{itemize}

\par Influence function-based (IF-based) methods estimate the effect of training data on model parameters using influence functions or related techniques, and perform unlearning via preconditioned gradient updates that negate this influence.

\begin{itemize} [leftmargin=0.5cm]
    \item \textbf{IDEA}~\cite{dong2024idea} is a certified unlearning framework that approximates sample influence to remove the forget set’s impact while preserving performance on retained data.
    \item \textbf{ScaleGUN}~\cite{yi2025scalable} improves scalability via randomized blockwise influence estimation, reducing the cost of inverse Hessian computation for large-scale graphs.
    \item \textbf{ETR}~\cite{yang2025erase} is a training-free two-stage approach. It first edits critical parameters associated with the data to forget, then applies a lightweight gradient approximation on the retained data to recover utility.
\end{itemize}

\begin{table*}[t]
    \centering
    \small
    \caption{Unlearning F1-score $\pm$ STD comparison(\%) under $10\%$ unlearning requests with SAGE backbone. 
    Results are reported on the Ele-Fashion and Goodreads-NC datasets across four multimodal encoder combinations. 
    A higher Unlearning F1 indicates better model utility on the test set.
    The best results in each setting are highlighted in \textcolor{red}{\textbf{red and bold}}.}
    \label{tab:mmgu_results_f1_combined}
    \renewcommand{\arraystretch}{1.2}
    \setlength{\tabcolsep}{4pt}
    \begin{tabular}{lcccccccc}
    \toprule
    \multirow{2}{*}{Method} & \multicolumn{4}{c}{Ele-Fashion} & \multicolumn{4}{c}{Goodreads-NC} \\
    \cmidrule(lr){2-5} \cmidrule(lr){6-9}
     & CLIP & T5+ViT & ImageBind & T5+DINOv2 & CLIP & T5+ViT & ImageBind & T5+DINOv2 \\
    \midrule
    \rowcolor{green!20} \multicolumn{9}{c}{\textbf{Node Unlearning}} \\
    BLPA & $84.24_{\pm 0.02}$ & $82.83_{\pm 0.10}$ & $84.08_{\pm 0.05}$ & $83.19_{\pm 0.05}$ & $74.14_{\pm 0.61}$ & $69.92_{\pm 0.74}$ & $64.56_{\pm 1.06}$ & $69.64_{\pm 0.51}$ \\
    BEKM & $85.09_{\pm 0.09}$ & $84.38_{\pm 0.08}$ & $86.56_{\pm 0.08}$ & $84.67_{\pm 0.07}$ & $78.03_{\pm 0.26}$ & $74.78_{\pm 0.32}$ & $70.64_{\pm 1.39}$ & $74.82_{\pm 1.04}$ \\
    GNNDelete & $60.76_{\pm 0.31}$ & $53.44_{\pm 0.15}$ & $70.32_{\pm 0.59}$ & $53.28_{\pm 1.55}$ & $62.43_{\pm 1.22}$ & $53.02_{\pm 0.10}$ & OOM & $52.13_{\pm 1.17}$ \\
    MEGU & $85.21_{\pm 0.06}$ & $83.80_{\pm 0.07}$ & $84.64_{\pm 0.49}$ & $83.42_{\pm 0.12}$ & $77.53_{\pm 0.06}$ & $74.93_{\pm 0.34}$ & $70.43_{\pm 0.13}$ & $74.56_{\pm 0.25}$ \\
    IDEA & $85.33_{\pm 0.04}$ & $84.34_{\pm 0.04}$ & $86.15_{\pm 0.01}$ & $84.49_{\pm 0.06}$ & $78.08_{\pm 0.17}$ & $73.21_{\pm 0.51}$ & $71.04_{\pm 1.47}$ & $75.68_{\pm 0.02}$ \\
    ScaleGUN & $78.21_{\pm 0.11}$ & $75.54_{\pm 0.43}$ & $79.96_{\pm 0.19}$ & $75.85_{\pm 0.21}$ & $66.07_{\pm 0.49}$ & $60.39_{\pm 0.00}$ & $62.08_{\pm 1.65}$ & $0.00_{\pm 0.00}$ \\
    ETR & $83.76_{\pm 0.02}$ & $82.75_{\pm 0.41}$ & $81.06_{\pm 0.85}$ & $81.17_{\pm 0.64}$ & $76.89_{\pm 0.45}$ & $68.24_{\pm 1.34}$ & $67.84_{\pm 0.36}$ & $71.84_{\pm 0.66}$ \\
    \textbf{FDQ} & \textcolor{red}{$\mathbf{86.92_{\pm 0.07}}$} & \textcolor{red}{$\mathbf{84.59_{\pm 0.13}}$} & \textcolor{red}{$\mathbf{86.69_{\pm 0.25}}$} & \textcolor{red}{$\mathbf{85.65_{\pm 0.29}}$} & \textcolor{red}{$\mathbf{82.68_{\pm 0.45}}$} & \textcolor{red}{$\mathbf{82.85_{\pm 0.28}}$} & \textcolor{red}{$\mathbf{79.92_{\pm 0.44}}$} & \textcolor{red}{$\mathbf{82.91_{\pm 0.02}}$} \\
    \midrule
    \rowcolor{yellow!20} \multicolumn{9}{c}{\textbf{Edge Unlearning}} \\
    BLPA & $84.48_{\pm 0.99}$ & $82.25_{\pm 0.31}$ & $84.80_{\pm 0.68}$ & $82.38_{\pm 0.67}$ & $70.95_{\pm 0.19}$ & $64.73_{\pm 0.37}$ & $63.24_{\pm 0.48}$ & $65.62_{\pm 0.30}$ \\
    BEKM & $83.91_{\pm 0.08}$ & $83.35_{\pm 0.29}$ & $86.10_{\pm 0.23}$ & $83.19_{\pm 0.18}$ & $75.16_{\pm 0.08}$ & $70.66_{\pm 0.03}$ & $66.49_{\pm 0.43}$ & $71.57_{\pm 0.30}$ \\ 
    GNNDelete & $64.22_{\pm 1.82}$ & $57.64_{\pm 1.43}$ & $69.40_{\pm 1.96}$ & $52.83_{\pm 0.67}$ & $54.37_{\pm 1.03}$ & $47.89_{\pm 2.17}$ & OOM & $55.48_{\pm 1.38}$ \\
    MEGU & $84.65_{\pm 0.14}$ & $81.34_{\pm 0.07}$ & $84.43_{\pm 0.37}$ & $82.32_{\pm 0.13}$ & $64.86_{\pm 0.35}$ & $56.79_{\pm 0.68}$ & $54.82_{\pm 1.30}$ & $57.51_{\pm 0.11}$ \\
    IDEA & $83.21_{\pm 0.02}$ & $82.77_{\pm 0.04}$ & $85.42_{\pm 0.32}$ & $82.80_{\pm 0.06}$ & $78.81_{\pm 0.09}$ & $77.53_{\pm 0.23}$ & $75.32_{\pm 0.62}$ & $77.22_{\pm 0.03}$ \\
    ScaleGUN & $78.30_{\pm 0.17}$ & $75.84_{\pm 0.31}$ & $80.06_{\pm 0.07}$ & $76.01_{\pm 0.13}$ & $65.02_{\pm 1.21}$ & $60.86_{\pm 0.78}$ & $60.39_{\pm 0.15}$ & $62.08_{\pm 0.93}$ \\
    ETR & $83.88_{\pm 0.03}$ & $83.95_{\pm 0.19}$ & $81.55_{\pm 0.83}$ & $82.88_{\pm 0.76}$ & $77.54_{\pm 0.56}$ & $74.02_{\pm 0.64}$ & $68.34_{\pm 0.87}$ & $69.41_{\pm 0.59}$ \\
    \textbf{FDQ} & \textcolor{red}{$\mathbf{86.62_{\pm 0.13}}$} & \textcolor{red}{$\mathbf{86.30_{\pm 0.39}}$} & \textcolor{red}{$\mathbf{86.94_{\pm 0.24}}$} & \textcolor{red}{$\mathbf{86.01_{\pm 0.46}}$} & \textcolor{red}{$\mathbf{82.88_{\pm 0.02}}$} & \textcolor{red}{$\mathbf{83.15_{\pm 0.02}}$} & \textcolor{red}{$\mathbf{80.28_{\pm 0.13}}$} & \textcolor{red}{$\mathbf{83.04_{\pm 0.05}}$} \\
    \bottomrule
    \end{tabular}
\end{table*}

\subsection{Experimental Setup}
\par All experiments are conducted on a Linux workstation equipped with an NVIDIA L40 GPU, using Python $3.10$, PyTorch $2.1.2$, and CUDA $12.1$. 
We adopt an $8/2$ train-test split and set the forgetting ratio to $10\%$ for both node and edge unlearning requests. 
We evaluate the robustness of our framework across GNN architectures, including GCN, GAT, and SAGE; SAGE achieves the best overall performance and is suitable for training on medium-to-large graphs due to its effective neighborhood aggregation. 
ScaleGUN is evaluated under its original implementation, where unlearning is performed on propagated node representations with a linear classifier, rather than via end-to-end updates of trainable GNN backbone parameters.
We further consider four multimodal encoding strategies for constructing node features. 
For aligned joint encoders, we use CLIP and ImageBind to jointly encode text and image into aligned embeddings. 
For separate encoders with concatenation, we encode textual content with T5 and encode visual content with either ViT or DINOv2, and then concatenate the two embeddings to obtain the final multimodal node features. 
Unless stated otherwise, we use the default hyperparameter settings for FDQ; a sensitivity study on $k$ and $\rho$ is reported in the Parameter Analysis section. 
To ensure statistical reliability, each experiment is repeated $10$ times with different random seeds, and we report the mean and standard deviation of the results.

\subsection{Comparison Results}
\par In this section, we compare FDQ with representative SOTA graph unlearning baselines under a $10\%$ forgetting ratio on Ele-Fashion and Goodreads-NC.
Unless otherwise noted, all baselines are discussed under their original implementation paradigms; in particular, ScaleGUN is evaluated under its original implementation protocol, where unlearning is performed on propagated node representations with a linear classifier, rather than via end-to-end updates of trainable GNN backbone parameters.
The evaluation is organized from four complementary perspectives: overall utility after unlearning, privacy-oriented forgetting effectiveness for node deletion, robustness under poisoned edge settings, and computational efficiency.
As shown in Tables~\ref{tab:mmgu_results_f1_combined}--\ref{tab:unlearning_time_sage_expanded}, FDQ consistently delivers a strong balance between utility preservation, forgetting quality, and runtime cost in multimodal graph scenarios.

\par Table~\ref{tab:mmgu_results_f1_combined} reports F1 performance after node and edge unlearning across four multimodal encoder combinations.
FDQ achieves the best utility in nearly all settings.
On Ele-Fashion, FDQ consistently outperforms strong baselines under both node and edge unlearning, with moderate but stable gains across encoder choices.
On Goodreads-NC, the advantages become larger: under node unlearning, FDQ remains the top method across all available encoder combinations, and under edge unlearning it again achieves the strongest overall performance.
In contrast, learning-based baselines such as GNNDelete exhibit notable utility degradation on multimodal graphs.
Moreover, on the large-scale Goodreads-NC dataset, GNNDelete encounters an OOM failure under ImageBind features, highlighting reduced scalability under high-dimensional multimodal inputs.
Compared with the smaller margins on Ele-Fashion, the larger gains on Goodreads-NC suggest that FDQ becomes increasingly beneficial as graph scale and multimodal complexity grow.
Overall, these results suggest that FDQ mitigates the utility degradation often observed in multimodal unlearning and supports the effectiveness of feature-dimension-aware parameter selection.

\par To further evaluate the effectiveness of node unlearning, we use MIA and report AUC-ROC in Figure~\ref{fig:mia_result}, where values closer to $50\%$ indicate stronger forgetting.
\begin{figure*}[t]
    \centering
    \includegraphics[width=0.95\linewidth]{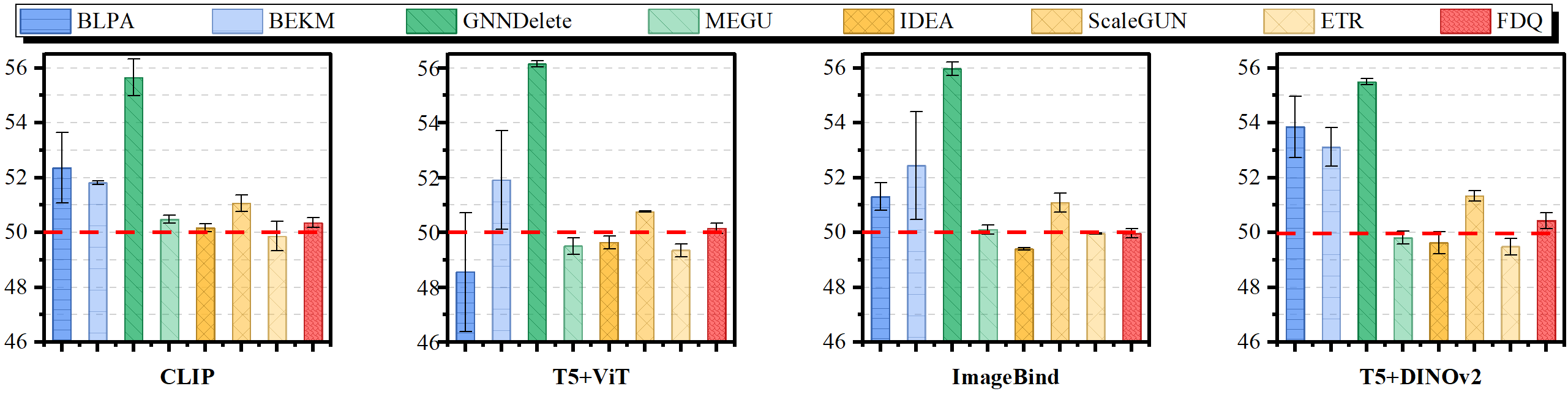}
    \caption{MIA results under node unlearning. AUC values closer to $50\%$ indicate stronger forgetting effectiveness.}
    \label{fig:mia_result}
\end{figure*}
The results show that FDQ keeps MIA AUC close to random-guessing across encoder settings, indicating that the attack model cannot reliably distinguish forgotten nodes from non-members.
In contrast, some baselines show consistently larger deviations from $50\%$; for example, GNNDelete exhibits systematically higher AUC values across encoder combinations, suggesting stronger residual membership signals.
Compared with these baselines, FDQ generally provides stronger privacy protection while still retaining competitive downstream utility.
These observations suggest that FDQ not only maintains accuracy after unlearning, but also effectively weakens membership signals associated with the forget set.

\par
\begin{figure*}[t]
    \centering
    \includegraphics[width=0.95\linewidth]{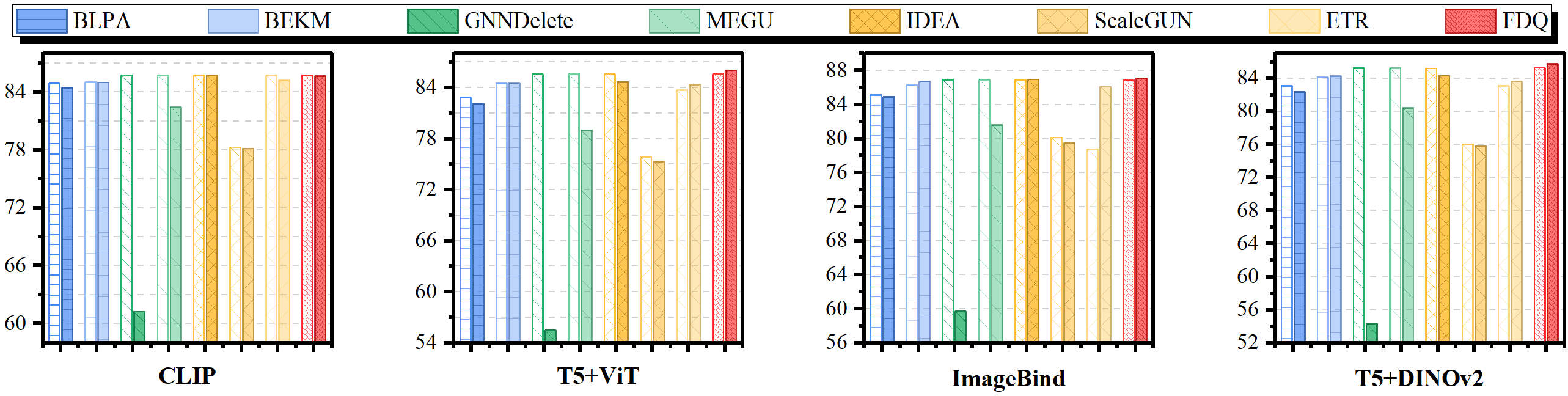}
    \caption{PA results under edge unlearning. Higher unlearning F1 indicates better removal of poisoned-edge influence while preserving utility.}
    \label{fig:pa_result}
\end{figure*}

 We then evaluate edge unlearning under poisoning attacks (PA), where the goal is to eliminate the influence of poisoned edges and recover robust task performance, and visualize the results in Figure~\ref{fig:pa_result}.
The PA figure compares poisoned-model F1 (light bars) and post-unlearning F1 (dark bars) for each method, and reveals clear differences in robustness across methods after edge unlearning.
GNNDelete shows a substantial drop from poisoned to unlearned performance across encoder settings, while MEGU also exhibits a milder but still noticeable decline.
In contrast, FDQ and strong baselines such as ETR generally maintain comparable, and in several settings slightly improved, unlearning F1 after removing poisoned edges.
These trends suggest that methods with more targeted parameter editing are better able to suppress poisoned-edge influence without excessively damaging utility-relevant knowledge.
Overall, FDQ delivers the most consistent high performance across multimodal encoder combinations, indicating strong suitability for edge-level forgetting under adversarial multimodal settings.

\par In addition, Table~\ref{tab:unlearning_time_sage_expanded} compares unlearning time on both datasets.
FDQ is significantly faster than most retraining-heavy or influence-estimation-heavy baselines, and remains sub-second on Ele-Fashion across all encoder settings.
On Goodreads-NC, FDQ is not the fastest but remains highly efficient, completing unlearning within a few seconds.
Nevertheless, FDQ offers a favorable efficiency-effectiveness trade-off when considered together with the stronger unlearning F1 results in Table~\ref{tab:mmgu_results_f1_combined} and Figure~\ref{fig:pa_result}.
These results suggest that FDQ maintains practical efficiency while preserving robust post-unlearning utility.
It is also worth noting that partition-based methods such as BLPA and BEKM typically require additional graph partitioning computation during model training.
In this work, the reported time only accounts for the unlearning process, without including the partition construction overhead.
Therefore, the end-to-end cost of these partition-based baselines is expected to be higher in practical settings.

\par Overall, the results across Tables~\ref{tab:mmgu_results_f1_combined}--\ref{tab:unlearning_time_sage_expanded} suggest that FDQ is a robust and efficient framework for multimodal graph unlearning.
FDQ preserves model utility, shows strong node and edge forgetting effectiveness under attack-based evaluation, and maintains low unlearning latency.
Taken together, these findings indicate that FDQ is a practical choice for privacy-aware lifecycle management of multimodal graph models.

\begin{table}[t]
    \centering
    \caption{Unlearning Time Performance on Ele-Fashion and Goodreads-NC (in seconds) with SAGE backbone.}
    \label{tab:unlearning_time_sage_expanded}
    \renewcommand{\arraystretch}{1.2}
    \setlength{\tabcolsep}{8pt}
    \begin{tabular}{lrrrr}
    \toprule
    \multirow{2}{*}{Method} & \multicolumn{4}{c}{Ele-Fashion} \\
    \cmidrule(lr){2-5}
     & CLIP & T5+ViT & ImageBind & T5+DINOv2 \\
    \midrule
    BLPA & $21.44$ & $26.56$ & $32.60$ & $26.82$ \\
    BEKM & $26.79$ & $34.78$ & $43.32$ & $76.03$ \\
    GNNDelete & $6.49$ & $8.71$ & $20.89$ & $8.56$ \\
    MEGU & $3.97$ & $5.44$ & $22.52$ & $5.38$ \\
    IDEA & $1.51$ & $1.85$ & $7.92$ & $1.83$ \\
    ScaleGUN & $37.24$ & $43.65$ & $46.09$ & $41.65$ \\
    ETR & $0.38$ & $0.75$ & $1.80$ & $1.29$ \\
    \textbf{FDQ} & $0.54$ & $0.66$ & $0.93$ & $0.70$ \\
    \midrule
    \multirow{2}{*}{Method} & \multicolumn{4}{c}{Goodreads-NC} \\
    \cmidrule(lr){2-5}
     & CLIP & T5+ViT & ImageBind & T5+DINOv2 \\
     \midrule
     BLPA & $134.94$ & $186.70$ & $231.86$ & $186.05$ \\
     BEKM & $133.55$ & $162.33$ & $185.10$ & $172.00$ \\
     GNNDelete & $18.54$ & $25.31$ & OOM & $25.30$ \\
     MEGU & $10.92$ & $15.42$ & $18.67$ & $15.38$ \\
     IDEA & $3.31$ & $5.71$ & $13.94$ & $5.72$ \\
     ScaleGUN & $64.60$ & $60.65$ & $86.71$ & $32.38$ \\
     ETR & $1.07$ & $1.54$ & $2.37$ & $1.62$ \\
     \textbf{FDQ} & $3.42$ & $3.61$ & $3.97$ & $3.64$ \\
     \bottomrule
    \end{tabular}
\end{table}

\subsection{Ablation Study}
\begin{table}[t]
    \centering
    \caption{Unlearning F1 $\pm$ STD (\%) ablation for FDQ and its variants under $10\%$ node unlearning.
    ``Ele'' denotes the Ele-Fashion, and ``Good" denotes the Goodreads-NC.}
    \label{tab:ablation_fdq}
    \renewcommand{\arraystretch}{1.2}
    \setlength{\tabcolsep}{5pt}
    \begin{tabular}{llccc}
        \toprule
        \textbf{Dataset} & \textbf{Encoder} & \textbf{FDQ} & \textbf{w/o FD} & \textbf{w/o Qtl.} \\
        \midrule
        \multirow{4}{*}{Ele}
            & CLIP & \textcolor{red}{$\mathbf{86.92_{\pm 0.07}}$} & $84.88_{\pm 1.08}$ & $81.84_{\pm 0.25}$ \\
            & T5+ViT & \textcolor{red}{$\mathbf{84.59_{\pm 0.13}}$} & $81.49_{\pm 1.16}$ & $79.49_{\pm 0.41}$ \\
            & ImageBind & \textcolor{red}{$\mathbf{86.69_{\pm 0.25}}$} & $83.76_{\pm 0.81}$ & $83.58_{\pm 0.09}$ \\
            & T5+DINOv2 & \textcolor{red}{$\mathbf{85.65_{\pm 0.29}}$} & $82.01_{\pm 1.64}$ & $82.72_{\pm 0.18}$ \\
        \midrule
        \multirow{4}{*}{Good}
            & CLIP & \textcolor{red}{$\mathbf{82.68_{\pm 0.45}}$} & $76.17_{\pm 1.35}$ & $73.44_{\pm 0.36}$ \\
            & T5+ViT & \textcolor{red}{$\mathbf{82.85_{\pm 0.28}}$} & $75.62_{\pm 1.54}$ & $69.99_{\pm 1.79}$ \\
            & ImageBind & \textcolor{red}{$\mathbf{79.92_{\pm 0.44}}$} & $70.28_{\pm 1.41}$ & $67.84_{\pm 2.66}$ \\
            & T5+DINOv2 & \textcolor{red}{$\mathbf{82.91_{\pm 0.02}}$} & $74.36_{\pm 0.53}$ & $72.19_{\pm 2.71}$ \\
        \bottomrule
    \end{tabular}
\end{table}

\par To assess key components of FDQ, we conduct an ablation study on two aspects: (i) the feature-dimension-aware quantile mechanism for layer-wise calibration under high-dimensional features, and (ii) the parameter selection rule mapping Fisher ratios to edit masks.
The study compares the full FDQ model against two variants:
(1)~w/o FD, which removes the feature-dimension-aware component and applies a uniform quantile $k$ across all layers on top of the same diagonal FIM computation, so input projections are no longer calibrated separately from deeper layers; and
(2)~w/o Qtl., which keeps the same FIM estimates but removes quantile-tail selection and instead applies a Fisher ratio-threshold test controlled by $\gamma$ (we set $\gamma{=}10$ following our main setup).
This variant parallels threshold-style, FIM-guided parameter selection in standard machine unlearning~\cite{foster2024fast}, which is orthogonal to the graph setting but motivates our ablation of quantile versus ratio-threshold masks under the same diagonal Fisher scores.
Experiments are conducted under a $10\%$ node unlearning scenario on Ele-Fashion and Goodreads-NC with the SAGE backbone and four multimodal encoder combinations (CLIP, T5+ViT, ImageBind, and T5+DINOv2).
We report Unlearning F1 on the test set as the utility metric in this ablation; we do not repeat MIA or wall-clock unlearning time here, since those are already compared against baselines in the previous subsections.

\par In all settings, full FDQ achieves the highest Unlearning F1, indicating that both dimension-aware quantile calibration and quantile-tail masking contribute to retaining utility after node unlearning.
Removing either component reduces F1; the drop is consistently larger on Goodreads-NC than on Ele-Fashion, which aligns with the higher structural and feature complexity of the larger graph.
Comparing the two ablated variants, w/o FD does not uniformly dominate w/o Qtl.: for example, on Ele-Fashion with T5+DINOv2, w/o Qtl. slightly exceeds w/o FD, whereas w/o FD is higher in the other seven settings.
Thus the advantage of FDQ over ratio-threshold selection is stable overall, while the relative ordering of uniform quantiles versus $\gamma$-thresholding depends on the encoder and dataset context.

\subsection{Parameter Analysis}
\par The hyperparameters $k$ and $\rho$ are crucial for FDQ's performance. 
$k$ determines how many parameters are selected for editing, thereby influencing the extent of forgetting and the post-unlearning utility. 
$\rho$ controls how conservatively we edit high-dimensional input-projection layers by tightening the effective selection quantile on those layers. 
In this section, we investigate their impact under the $10\%$ node unlearning setup with the SAGE backbone. 
We report Unlearning F1 on the test set as the utility metric. 
Unless specified otherwise, we keep $\tau$ and $k_{\min}$ fixed to the default values in our main configuration.

\begin{figure}[t]
    \centering
    \includegraphics[width=0.95\linewidth]{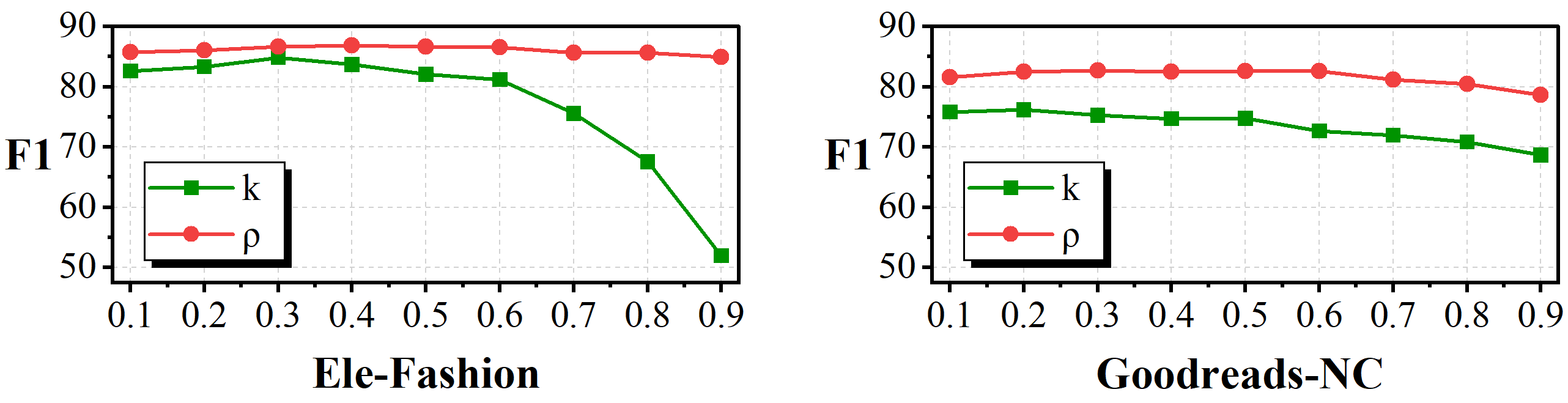}
    \caption{Parameter analysis of $k$ and $\rho$ on Ele-Fashion and Goodreads-NC (Unlearning F1, \%).}
    \label{fig:param_analysis}
\end{figure}

\par We investigate the effect of different $k$ values by setting $\rho{=}1$ and sweeping $k \in [0.1, 0.9]$ with a step size of $0.1$, which recovers a uniform-quantile baseline across layers. 
On Ele-Fashion, increasing $k$ improves Unlearning F1 up to a moderate range (peaking at $k{=}0.3$), but performance declines when $k$ becomes too large. 
On Goodreads-NC, the best result is achieved at a slightly smaller value ($k{=}0.2$), after which F1 gradually decreases as $k$ increases. 
This is because a larger $k$ selects more parameters for editing, which can strengthen forgetting but may also over-edit and remove useful knowledge. 
Thus we set $k^*{=}0.3$ for Ele-Fashion and $k^*{=}0.2$ for Goodreads-NC (Fig.~\ref{fig:param_analysis}), and use the corresponding $k^*$ when tuning $\rho$ on each dataset.

\par We tune $\rho$ to investigate how feature-dimension-aware tightening affects performance. 
With $k$ fixed to each dataset's $k^*$, we sweep $\rho \in [0.1, 0.9]$ with a step size of $0.1$. 
We observe that moderate $\rho$ values yield the best utility (Ele-Fashion peaks at $\rho{=}0.4$, while Goodreads-NC remains near-optimal for $\rho \in [0.2, 0.6]$), and that overly small or large $\rho$ leads to suboptimal performance. 
Intuitively, smaller $\rho$ enforces more conservative edits on high-dimensional input-projection layers, which helps preserve multimodal representations, whereas excessively small $\rho$ can make edits overly conservative. 
Conversely, very large $\rho$ reduces the benefit of feature-dimension-aware tightening and may expose wide input projections to stronger edits. 
Overall, $k$ controls the overall edit strength, while $\rho$ calibrates how aggressively FDQ protects high-dimensional input projections.



\section{Conclusion}
In this paper, we propose FDQ, a feature-dimension-aware quantile framework for multimodal graph unlearning. Existing parameter-editing methods apply uniform quantile selection across layers, which can over-edit sensitive input projections in high-dimensional multimodal graphs and cause severe utility degradation. FDQ mitigates this by identifying high-dimensional input layers and applying more conservative, layer-wise quantile thresholds, while retaining diagonal FIM-based importance estimation. This yields a unified and efficient parameter-dampening pipeline for both node and edge unlearning without retraining. Experiments on Ele-Fashion and Goodreads-NC show that FDQ consistently preserves utility, maintains strong resistance to membership inference attacks, and improves robustness under poisoned-edge settings, while achieving low latency. A limitation is its sensitivity to hyperparameters controlling edit strength and feature-dimension-aware calibration, which may require tuning in practice.

\bibliographystyle{IEEEtran}
\bibliography{ref}

\vspace{-12pt}
\begin{IEEEbiography}[{\includegraphics[width=1in,height=1.25in,clip,keepaspectratio]{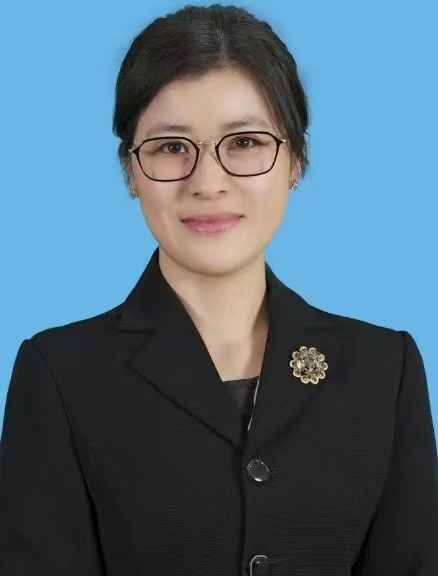}}] {Jingjing Zhou} received the M.S. degree from Shandong University in 2004 and the Ph.D. degree from University of Science and Technology of Beijing in 2009. 
She is currently Associate Professor in School of Information and Electronic Engineering, Zhejiang Gongshang University, China. 
Her research interests include artificial intelligence, graph learning, and computer networks.
\end{IEEEbiography}

\vspace{-12pt}
\begin{IEEEbiography}[{\includegraphics[width=1in,height=1.25in,clip,keepaspectratio]{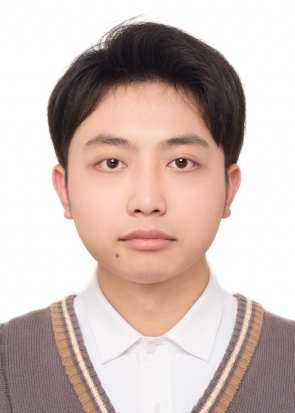}}] {Yongshuai Yang} is an M.S. student at School of Information and Electronic Engineering, Zhejiang Gongshang University, China. 
He received the bachelor's degree from Zhejiang Gongshang University in 2024. 
His research focuses on artificial intelligence and graph learning.
\end{IEEEbiography}

\vspace{-12pt}
\begin{IEEEbiography}[{\includegraphics[width=1in,height=1.25in,clip,keepaspectratio]{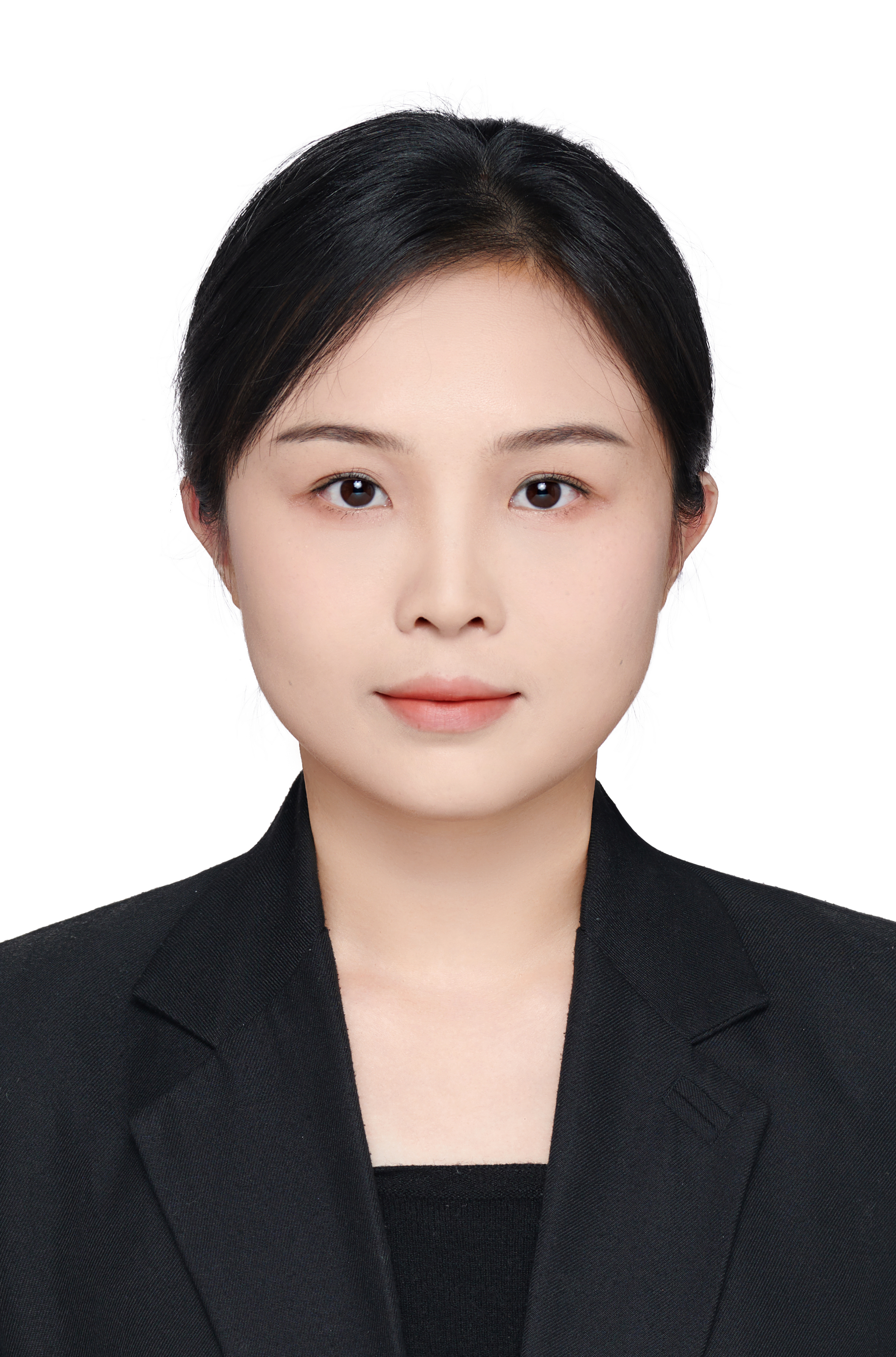}}] {Qing Qing} is currently a PhD student in College of Computer Science and Technology, Jilin University, Changchun, China.
Before that, she received the B.Sc. degree from Northeast Agricultural University, Harbin, China, in 2018, and the M.Sc. degree from Dalian University of Technology, Dalian, China, in 2021. 
Her research interests include graph learning, algorithmic fairness, responsible AI.
\end{IEEEbiography}

\vspace{-12pt}
\begin{IEEEbiography}[{\includegraphics[width=1in,height=1.25in,clip,keepaspectratio]{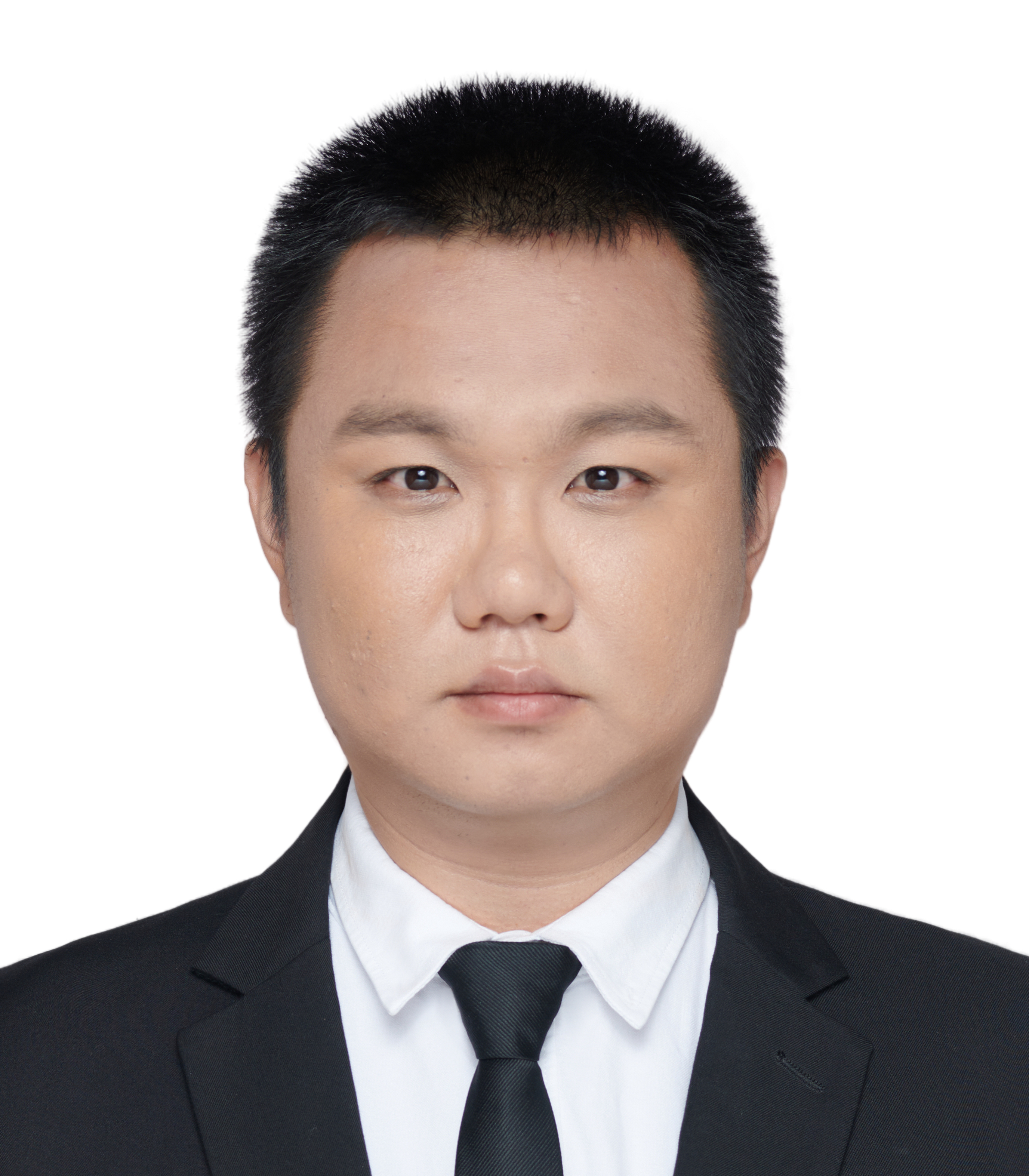}}]{Renqiang Luo}
  received the B.Sc. degree from University of Science and Technology of China, Hefei, China, in 2016, and the M.Sc. degree from University of South Australia, Adelaide, Australia, in 2019. He received a Ph.D. degree in the School of Software, Dalian University of Technology, Dalian, China, in 2024. Dr. Renqiang Luo is currently an Assistant Professor in the Jilin University, Changchun, China. His research interests include graph learning, algorithmic fairness, and trustworthy AI.
\end{IEEEbiography}

\vspace{-12pt}
\begin{IEEEbiography}[{\includegraphics[width=1in,height=1.25in,clip,keepaspectratio]{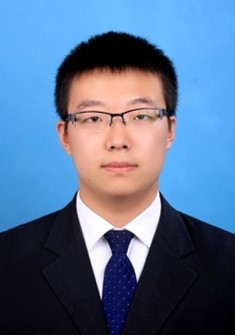}}]{Ziqi Xu} received the M.S. degree in Computing and Innovation from the School of Computer and Mathematical Sciences, The University of Adelaide, Australia, and the Ph.D. degree in Computer Science from the University of South Australia, Australia. 
He is currently a Lecturer in Data Science and Artificial Intelligence with the School of Computing Technologies, RMIT University, Australia. 
His research interests include responsible AI, causal inference, fairness, and explainable machine learning.
\end{IEEEbiography}

\vspace{-12pt}
\begin{IEEEbiography}[{\includegraphics[width=1in,height=1.25in,clip,keepaspectratio]{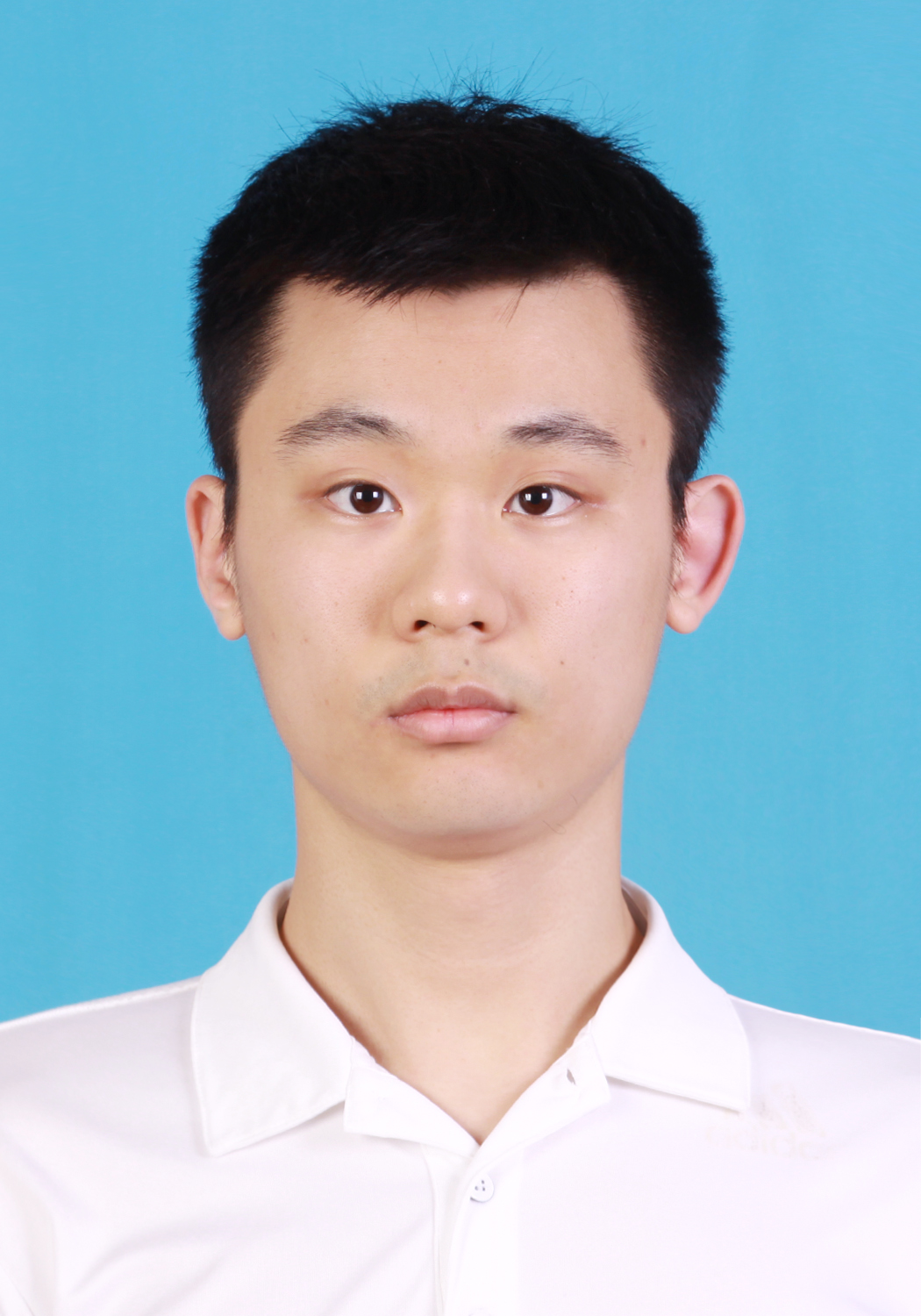}}]{Xikun Zhang} is a Lecturer at the School of Computing Technologies at RMIT University. He received his Ph.D. from the School of Computer Science at the University of Sydney. His research interests span deep graph learning, reasoning with large language models, and biomedical AI. His work has been published in leading conferences and journals, including ICLR, NeurIPS, KDD, ICDM, CVPR, ECCV, TPAMI, and TNNLS.
\end{IEEEbiography}

\vspace{-12pt}
\begin{IEEEbiography}[{\includegraphics[width=1in,height=1.25in,clip,keepaspectratio]{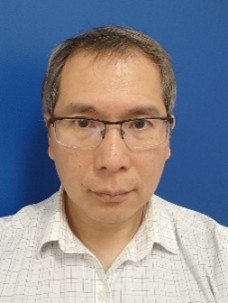}}]
{Ivan Lee} (Senior Member, IEEE) received BEng, MCom, MER, and PhD degrees from the University of Sydney. He is a Professor and Academic Lead: International and Engagement, School of Computer Science and IT, Adelaide University. Prior to this appointment, he worked at University of South Australia, Ryerson University, Remotek Corporation, and Cisco Systems. He was a REDI Fellow in 2023. He currently serves as the Program 2 Co-Lead of the ARC Research Hub for Intelligent Contaminant-Sensing in complex Environments (IC-SensE Hub), and as an Associate Editor of IEEE Transactions of Multimedia. His research interests include intelligent sensors, multimedia system, and data science.
\end{IEEEbiography}

\vspace{-12pt}
\begin{IEEEbiography}[{\includegraphics[width=1in,height=1.25in,clip,keepaspectratio]{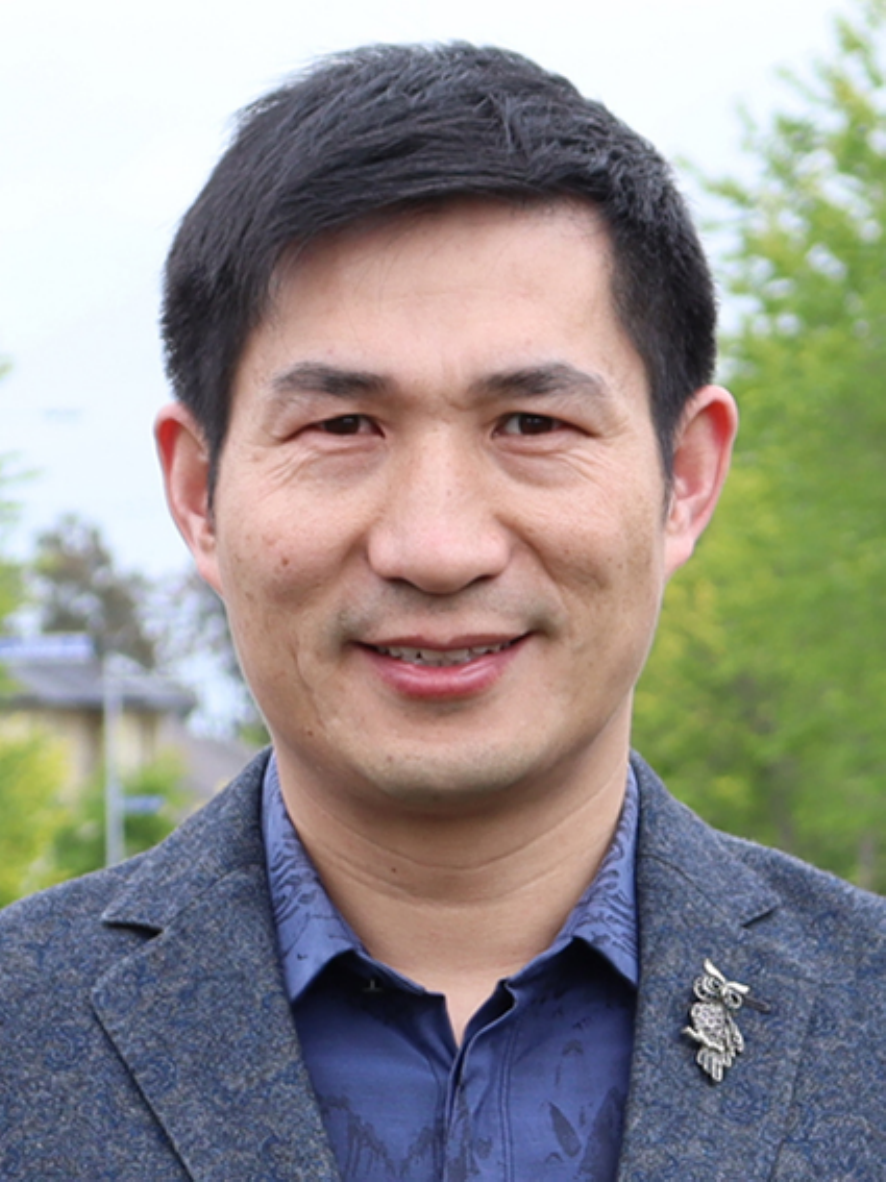}}]{Feng Xia}
(Fellow, IEEE) received the BSc and PhD degrees from Zhejiang University, Hangzhou, China. He is a Professor in School of Computing Technologies, RMIT University, Australia. Recognized as a Clarivate Highly Cited Researcher and a ScholarGPS Highly Ranked Scholar, Dr. Xia has published over 400 scientific papers. His work is featured in top-tier journals and conferences. Dr. Xia has extensive editorial and organizational experience, having served as an Associate or Guest Editor for over 20 journals and in various Chair roles for more than 30 conferences. His contributions and leadership have been recognized by prestigious awards. He has delivered numerous keynote speeches and invited talks at international venues worldwide. He is the Chair of IEEE Task Force on Learning for Graphs. His research interests include artificial intelligence, graph learning, brain, robotics, and cyber-physical systems. He is a Fellow of the IEEE.
\end{IEEEbiography}

\vfill
\end{document}